%% file: main.tex
\documentclass[10pt,twocolumn,letterpaper]{article}

\usepackage{cvpr}              % To produce the CAMERA-READY version
\usepackage{textcomp}
\usepackage{amsmath}
\usepackage{multirow}
\usepackage{colortbl}
\usepackage{pifont}
\usepackage{xfrac}

\definecolor{cvprblue}{rgb}{0.21,0.49,0.74}
\definecolor{mblue}{rgb}{0.00, 0.44, 0.75}
\definecolor{mred}{rgb}{0.75, 0.00, 0.00}
\usepackage[pagebackref,breaklinks,colorlinks,allcolors=cvprblue]{hyperref}

\def\paperID{10331} % *** Enter the Paper ID here
\def\confName{CVPR}
\def\confYear{2025}

\title{ReferDINO-Plus: 2nd Solution for 4th PVUW MeViS Challenge at CVPR 2025}

\author{
Tianming Liang \quad Haichao Jiang \quad Wei-Shi Zheng \quad Jian-Fang Hu\thanks{Corresponding author.} \\[2ex] 
Sun Yat-sen University  
}
\begin{document}
\maketitle
\begin{abstract}
    Referring video object segmentation (RVOS) aims to segment target objects throughout a video based on a text description. This task has attracted increasing attention in the field of computer vision due to its promising applications in video editing and human-agent interaction.
    Recently, ReferDINO showcases promising performance in this task by adapting the object-level vision-language knowledge from pretrained foundational image models.
    In this report, we extend its capabilities by incorporating SAM2 to enhance mask quality and object consistency.
    To effectively balance performance between single-object and multi-object scenarios, we introduce a conditional mask fusion strategy that adaptively combines masks from ReferDINO and SAM2. Our solution, termed ReferDINO-Plus, achieves 60.43 \(\mathcal{J}\&\mathcal{F}\) on MeViS test set, securing 2nd place in the MeViS PVUW challenge at CVPR 2025. The code is available at: \url{https://github.com/iSEE-Laboratory/ReferDINO-Plus}.
\end{abstract}

\section{Introduction}
\label{sec:intro}
Referring video object segmentation (RVOS) aims to segment target objects throughout a video based on a text description. 
This task bridges the gap between vision-language understanding and pixel-level video analysis, offering significant value for many down-stream applications, such as video editing and human-agent interaction systems.

Previous RVOS datasets like Refer-Youtube-VOS~\cite{seo2020urvos} and Ref-DAVIS17~\cite{khoreva2019video} focused on segmenting salient video objects described by static attributes (e.g., color and shape) or simple spatial relationships, overlooking the complex, dynamic properties in real-world scenarios. 
To encourage more efforts towards challenging yet practical pixel-level video understanding, the 4th PVUW workshop at CVPR 2025 presents a challenging RVOS benchmark MeViS~\cite{ding2023mevis} for competition.
Different from previous RVOS datasets, MeViS~\cite{ding2023mevis} focuses on the understanding of temporal motion in the RVOS task. In MeViS, the videos often contain multiple objects with similar static appearances but different motion attributes, and the object descriptions in MeViS mainly focus on motion and temporal expressions. 
In addition, MeViS includes numerous multi-object expressions, allowing for the referral of an unlimited number of target objects in the video.
These features make MeViS more challenging and reflective of real-world scenarios.
To overcome these challenges, a strong cross-modal spatiotemporal capability is necessary to understand the motion properties in the videos and descriptions. 

Early works~\cite{bellver2010refvos} in RVOS tend to directly apply the referring image segmentation methods~\cite{ding2021vision,yang2022lavt} to RVOS. 
However, this manner ignores the temporal information and often result in inconsistent object prediction.
Afterwards, MTTR~\cite{botach2022end} introduced the DETR paradigm~\cite{carion2020end} into RVOS. Building on this, ReferFormer~\cite{wu2022language} proposed to generate queries directly from the text description. Subsequent works~\cite{han2023html, miao2023spectrum, luo2023soc, yuan2024losh} have focused on modular improvements to enhance cross-frame consistency and temporal understanding.
Despite these efforts, current RVOS models~\cite{luo2023soc,he2024decoupling,yan2024referred} still struggle with insufficient vision-language understanding, often failing to handle complicated object descriptions, especially involving composite appearance, location and attributes.
Recently, ReferDINO~\cite{liang2025referdino} was proposed to address this limitation by leveraging the pretrained vision-language knowledge from the foundational visual-grounding model GroundingDINO~\cite{liu2024grounding}.
To enable end-to-end adaptation on RVOS data, ReferDINO incorpoarates a cross-modal temporal enhancer and a well-designed mask decoder.
Combining these components, ReferDINO achieves state-of-the-art performance across various RVOS benchmarks.

% Meanwhile, the development of SAM2 ..

Our solution, termed \textbf{ReferDINO-Plus}, is a two-stage strategy built upon ReferDINO and SAM2~\cite{ravi2024sam2}. 
Specifically, in the first stage, we employ ReferDINO to perform cross-modal object identification and spatiotemporal dense reasoning. Given a video and an object description, ReferDINO generates masks and binary scores for each candidate target. However, due to the lack of training on large-scale segmentation data, the mask quality may be unsatisfactory.
Therefore, in the second stage, we apply SAM2 for mask refinement and augmentation, regarding the frame and mask with the highest binary score as the prompts.
After this two-stage process, we can obtain two series of masks---one from ReferDINO and the other from SAM2. 
Intuitively, the masks from SAM2 are more reliable and stable. However, we observe that SAM2 tends to degenerate the multi-object mask into a single-object mask, resulting in performance degradation in multi-object scenarios.
To address this issue, we design a Conditional Mask Fusion (CMF) strategy. For single-object cases, we output only the masks from SAM2; for multi-object cases, we combine both the masks from ReferDINO and SAM2. 
However, it remains challenging to determine whether an expression involves multiple objects. 
In our experiment, we define it as a multi-object case if the mask area of SAM2 is less than $2/3$ of ReferDINO's.
Our solution is straight-forward yet effective.
Without further finetuning with additional pseudo labels on validation/test data~\cite{fang2024uninext}, our solution achieves 55.27 \(\mathcal{J}\&\mathcal{F}\) on MeViS validation set, and 60.43 \(\mathcal{J}\&\mathcal{F}\) on MeViS test set, securing the final ranking of 2nd in the MeViS Track at CVPR 2025 PVUW challenge.

\section{Related Works}
\label{sec:relate_works}
\subsection{Referring Video Object Segmentation} 
RVOS~\cite{gavrilyuk2018actor,ding2023mevis,seo2020urvos} aims to segment objects throughout the video based on text descriptions.
Some works~\cite{bellver2010refvos} attempt to directly apply the referring image segmentation methods~\cite{ding2021vision,yang2022lavt} to RVOS. 
However, this manner is unable to capture temporal information and often result in inconsistent object prediction.
MTTR~\cite{botach2022end} firstly introduces the DETR paradigm~\cite{carion2020end} into RVOS. Furthermore, ReferFormer~\cite{wu2022language} proposes to produce the queries from the text description.
On the top of this pipeline, follow-up works~\cite{han2023html,miao2023spectrum,luo2023soc,yuan2024losh} focus on modular improvements to improve cross-frame consistency and temporal understanding.
For example, SOC~\cite{luo2023soc} aggregates video content and textual guidance with a semantic integration module for unified temporal modeling and cross-modal alignment.
DsHmp~\cite{he2024decoupling} decouples video-level referring expression understanding into static and motion perception, with a customized module to enhance temporal comprehension.
Despite notable progress on specific datasets, these models are limited by insufficient vision-language understanding, and often struggle in unseen objects or scenarios.
Recently, ReferDINO~\cite{liang2025referdino} overcomes this limitation by leveraging the pretrained vision-language knowledge from GroundingDINO~\cite{liu2024grounding}, and extending the capabilities of dense perception and spatio-temporal reasoning by integrating an effective mask decoder and a temporal enhancer. 

\subsection{Semi-supervised Video Object Segmentation} 
The conventional semi-supervised video object segmentation aims to propagate the ground-truth object masks from a given frame throughout the video. Many existing approaches~\cite{cheng2022xmem,cheng2024putting,ravi2024sam2} employ a memory mechanism to store the past features for tracking and segmenting on future frames. 
Early deep learning-based methods mainly employed online adaptation strategies, where models were fine-tuned either on the initial frame or all frames to specialize in the target object.
To reduce the computation overheads, many works focus on offline training conditioned solely on the first frame or incorporating temporal dependencies from preceding frames.
With the development of the strong Segment Anything Model (SAM)~\cite{kirillov2023segment}, many efforts attempt to combine SAM with video trackers based on masks to perform segmentation throughout the video. 
However, this manner is ineffective since the combination is not end-to-end differentiable.
To address this limitation, SAM2~\cite{ravi2024sam2} was proposed, aiming to accommodate the specific demand of semi-supervised video object segmentation.
SAM2 showcases strong performance in object tracking and segmentation, and has attracted more and more attention in the computer vision field.

\begin{figure*}[t]
    \centering
    \includegraphics[scale=0.32]{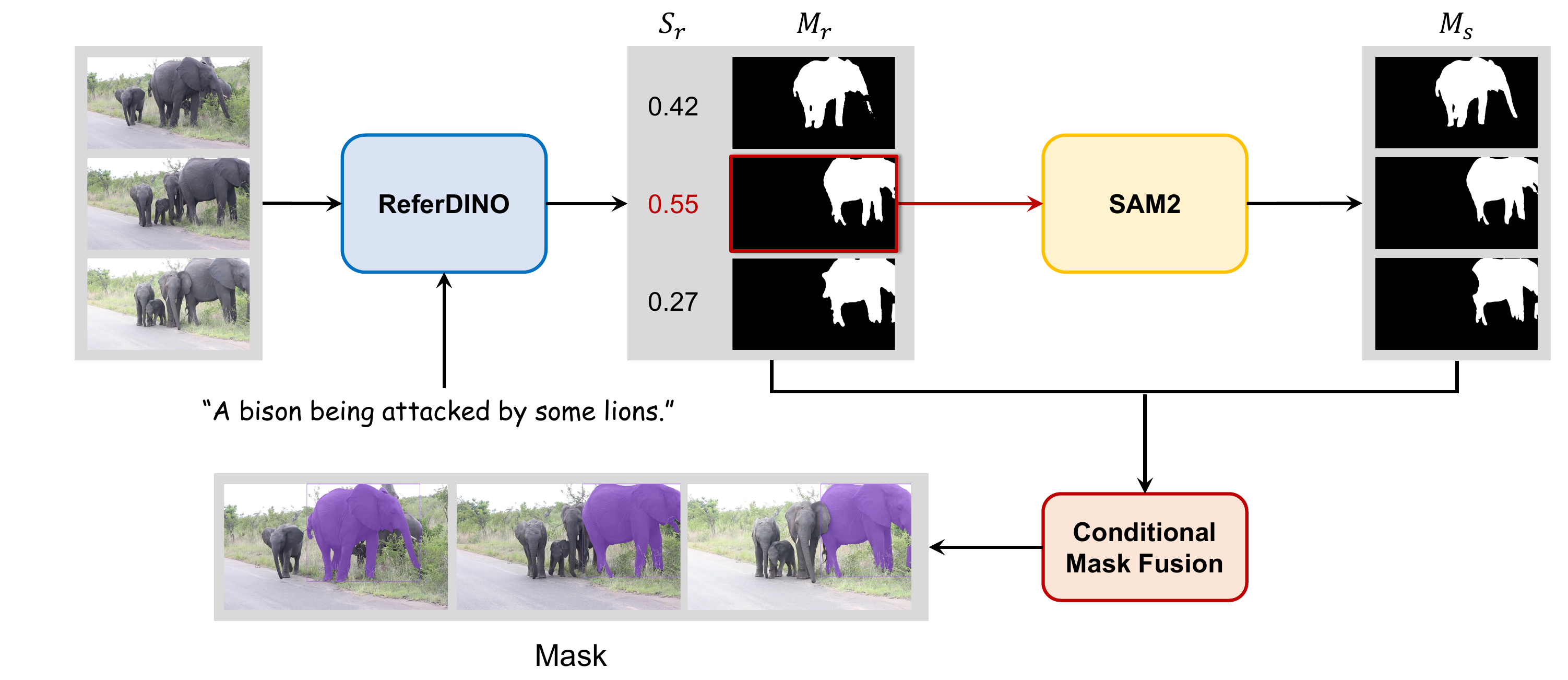}
    \caption{Overview of our solution \textbf{ReferDINO-Plus}. For each video-description pair, we input it into ReferDINO to derive the object masks $M_r$ and the corresponding scores $S_r$ across the frames. Then, we select the mask with highest score as the prompt of SAM2, producing refined masks $M_s$. Finally, we fuse the two series of masks through the \textit{conditional mask fusion} strategy.
    Best view in color.}
    \label{fig:model}
 \end{figure*}

\section{ReferDINO-Plus}
The overall framework of our solution \textbf{ReferDINO-Plus} is presented in Figure~\ref{fig:model}. For each video-description pair, we input it into ReferDINO to derive the object masks and the corresponding scores across the frames. Then, we select the mask with highest score as the prompt of SAM2, producing refined masks. Finally, we fuse the two series of masks through the conditional mask fusion strategy, to generate the final masks on each frame.

\subsection{Cross-modal Reasoning with ReferDINO}
ReferDINO~\cite{liang2025referdino} is a strong RVOS model that inherits object-level vision-language knowledge from GroundingDINO~\cite{liu2024grounding}, and is further endowed with pixel-level dense prediction and cross-modal spatiotemporal reasoning. Formally, given a video clip of $T$ frames and a text description, ReferDINO performs cross-modal reasoning and segmentation, deriving a mask sequence $\{M_r^t\}_{t=1}^T$ and the corresponding scores $\{S_r^t\}_{t=1}^T$ throughout the video. Following the practice in previous works~\cite{liang2025referdino,ding2023mevis,he2024decoupling}, we combine the multiple object masks with scores higher than a preset threshold $\sigma$ to handle the multi-object cases.

\subsection{Mask Refinement with SAM2}
SAM2~\cite{ravi2024sam2} is a strong prompt-based segmentation model, which can efficiently produce high-quality object masks throughout the video based on the given prompts, in form of clicks, bounding boxes and masks.
In this work, we utilize SAM2 to further refine the mask quality and object consistency of ReferDINO.
Specifically, after obtaining the masks and corresponding scores across the frames, we select the mask with the maximum score as the prompt. Based on the prompt frame and mask, SAM2 produces a refined mask sequence $\{M_s^t\}_{t=1}^T$ throughout the video.

\subsection{Conditional Mask Fusion}
Although the masks from SAM2 are more reliable and stable, we observe that SAM2's overall performance on MeViS is significantly weaker than that of ReferDINO. In our experiments, we identify the main reason as that, for multi-object mask prompts, SAM2 tends to degenerate them into single-object masks, leading to a substantial target loss in subsequent frames.
To address this issue, we design a \textit{Conditional Mask Fusion} (CMF) principle: for single-object cases, we output only the masks from SAM2; for multi-object cases, we combine both the masks from ReferDINO and SAM2. 

However, it remains challenging to determine whether an expression involves multiple objects. 
In our solution, we define it as a multi-object case if the mask area of SAM2 is less than $2/3$ of ReferDINO's. 
Formally, this process can be described as follows:
\begin{equation}
    M = 
    \begin{cases}
    M_s + M_r & \text{if} \  \mathcal{A}(M_s) < \frac{2}{3} \mathcal{A}(M_r) \\
    M_s & \text{otherwise}
    \end{cases}
\end{equation}
where $\mathcal{A}(\cdot)$ indicates the mask area. 
Note that our CMF is conducted individually on each frame, which empirically achieves better performance. 

\section{Experiment}
\subsection{Dataset and Metrics}
MeViS~\cite{ding2023mevis} is a large RVOS dataset comprising 2K videos and 28K text descriptions.
In this competition, the provided test set includes 100 videos and 1,456 language descriptions. These descriptions may correspond to a single object, multiple objects, or even non-objects within the videos, making the dataset significantly challenging.
We employ region similarity $\mathcal{J}$ (average IoU), contour accuracy F (mean boundary similarity), and their average $\mathcal{J}\&\mathcal{F}$ as the evaluation metrics.

\subsection{Implementation Details}
We pretrain ReferDINO~\cite{liang2025referdino} on the referring image segmentation datasets RefCOCO/+/g~\cite{kazemzadeh2014referitgame,mao2016generation} at first, and then train with the combination of Refer-Youtube-VOS~\cite{seo2020urvos} and Ref-DAVIS17~\cite{khoreva2019video}. Finally, we finetune it on the training set of MeViS.
For ReferDINO, we use the MM-GroundingDINO-SwinB~\cite{zhao2024open} as the backbone.
For SAM2, we use the Sam2.1\_Hiera\_Large as the backbone.
We set the threshold $\sigma=0.275$, other hyper-parameters are the same as the original ReferDINO.
Unlike the solutions~\cite{fang2024uninext} in previous challenges, we do not use additional pseudo labels on the validation or test data to for further finetuning.

\begin{table}[t]
    \centering
    \setlength{\tabcolsep}{4mm} % Adjusted for better spacing
    {\begin{tabular}{l|ccc}
        \toprule
        Team & $\mathcal{J}$\&$\mathcal{F}$ & $\mathcal{J}$ & $\mathcal{F}$ \\
        \midrule
        MVP-Lab & 61.98 & 58.83 & 65.14 \\
        \textbf{ReferDINO-Plus} & 60.43 & 56.79 & 64.07 \\
        HarborY & 56.26 & 52.68 & 59.84 \\
        Pengsong & 55.91 & 53.06 & 58.76 \\
        ssam2s & 55.16 & 52.00 & 58.33 \\
        strong\_kimchi & 55.02 & 51.78 & 58.27 \\
        \bottomrule
    \end{tabular}}
    \caption{The leaderboard of the MeViS test set.}
    \label{tab:main}
\end{table}

\begin{figure*}[t]
    \centering
    \includegraphics[scale=0.48]{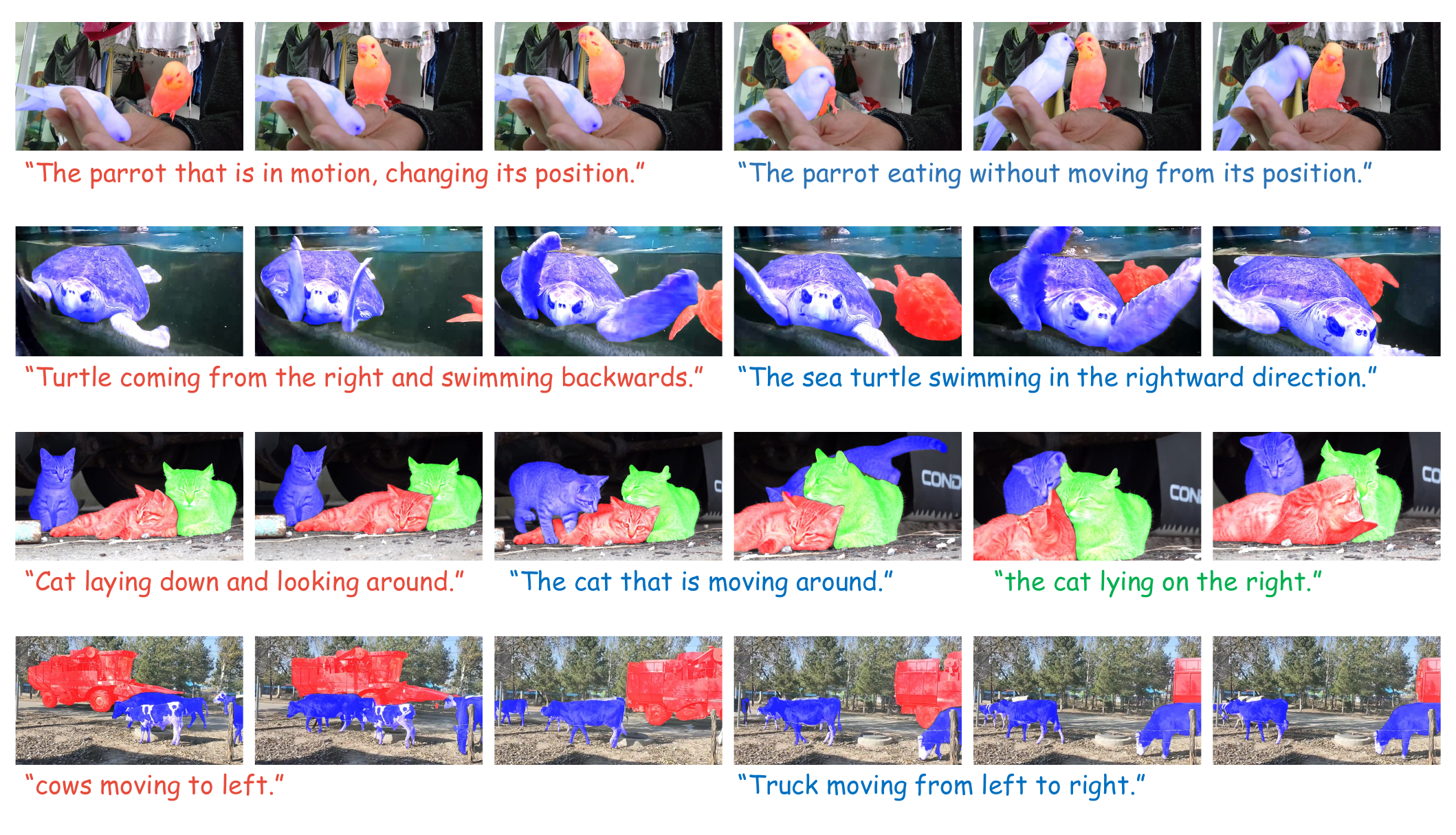}\vspace{-1em}
    \caption{Visualization results of our solution on MeViS test set.}
    \label{fig:visual}
 \end{figure*}

\subsection{Competition Results}
As shown in Table~\ref{tab:main}, our solution achieves 60.43 \(\mathcal{J}\&\mathcal{F}\) on MeViS test set, securing the final ranking of 2nd in the MeViS Track at CVPR 2025 PVUW challenge.

\begin{table}[t]
    \centering
    \setlength{\tabcolsep}{4mm} % Adjusted for better spacing
    {\begin{tabular}{l|ccc}
        \toprule
        Method & $\mathcal{J}$\&$\mathcal{F}$ & $\mathcal{J}$ & $\mathcal{F}$ \\
        \midrule
        ReferDINO & 51.67 & 47.94 & 55.40 \\
        +SAM2 & 52.54 & 49.18 & 55.90 \\
        +SAM2+CMF$_V$ & 54.82 & 51.39 & 58.24 \\
        +SAM2+CMF & 55.27 & 51.80 & 58.75 \\
        \bottomrule
    \end{tabular}}
    \caption{Ablation stuides on the MeViS validation set.}
    \label{tab:ab}
\end{table}

\subsection{Ablation Studies}
We conduct abaltion studies on MeViS validation set to explore the effects of individual components in our solution.
As shown in Table~\ref{tab:ab}, the refinement of SAM2 improves 3.15\% \(\mathcal{J}\&\mathcal{F}\).
When we perform CMF on the entire video (termed CMF$_V$), the result can be improved by 2.28\% \(\mathcal{J}\&\mathcal{F}\). 
When performing CMF on individual frames, the result can be further improved by 0.45\%. These results demonstrate the effectiveness of our components.

\subsection{Visualization}
In Figure~\ref{fig:visual}, we present several visualization results of ReferDINO-Plus on MeViS test set. It shows that our method can effectively segment the targets based on the corresponding text descriptions throughout the videos. These results demonstrate the accurate and high-quality masks generated from our ReferDINO-Plus.
In the 4th line of Figure~\ref{fig:visual}, we further show a case of mult-object referring, which demonstrates the effectiveness of our conditional mask fusion strategy.

\section{Conclusion}
In this work, we propose ReferDINO-Plus to address the problem of motion expression guided video object segmentation. 
This is a two-stage strategy. In the first stage, it employs ReferDINO to perform cross-modal object identification and spatiotemporal dense reasoning. In the second stage, it integrates SAM2 for mask refinement and object tracking. 
To address the multi-object collapse problem, we further design a conditional mask fusion strategy for post segmentation ensemble. Without further finetuning with additional pseudo labels on validation/test data, our solution achieves the 2nd place in PVUW MeViS challenge at CVPR 2025.
{
    \small
    \bibliographystyle{ieeenat_fullname}
    \bibliography{main}
}


\end{document}